\documentclass[10pt,twocolumn,letterpaper]{article}

\usepackage[pagenumbers]{cvpr} % To force page numbers, e.g. for an arXiv version
\pdfoutput=1
\usepackage{graphicx}
\usepackage{amsmath}
\usepackage{amssymb}
\usepackage{booktabs}
\usepackage{multirow}
\usepackage{color,xcolor}
\usepackage{bm}
\usepackage{pifont}% http://ctan.org/pkg/pifont
\usepackage[accsupp]{axessibility}
\usepackage{colortbl}
\usepackage{listings}
\usepackage{amsthm}
\usepackage{tabulary}
\usepackage{colortbl}
\usepackage{enumitem}
\usepackage{braket}
\usepackage{array}
\usepackage{makecell}
\usepackage{float}
\usepackage{algorithm}
\usepackage{bbding}
\usepackage{tabu}
\usepackage{pifont} 
\usepackage[accsupp]{axessibility}
\newcommand{\cmark}{\ding{51}}%
\newcommand{\xmark}{\ding{55}}%

\def\x{\times}

\newlength\savewidth\newcommand\shline{\noalign{\global\savewidth\arrayrulewidth
  \global\arrayrulewidth 1pt}\hline\noalign{\global\arrayrulewidth\savewidth}}

\makeatletter
\def\UrlAlphabet{%
      \do\a\do\b\do\c\do\d\do\e\do\f\do\g\do\h\do\i\do\j%
      \do\k\do\l\do\m\do\n\do\o\do\p\do\q\do\r\do\s\do\t%
      \do\u\do\v\do\w\do\x\do\y\do\z\do\A\do\B\do\C\do\D%
      \do\E\do\F\do\G\do\H\do\I\do\J\do\K\do\L\do\M\do\N%
      \do\O\do\P\do\Q\do\R\do\S\do\T\do\U\do\V\do\W\do\X%
      \do\Y\do\Z}
\def\UrlDigits{\do\1\do\2\do\3\do\4\do\5\do\6\do\7\do\8\do\9\do\0}
\g@addto@macro{\UrlBreaks}{\UrlOrds}
\g@addto@macro{\UrlBreaks}{\UrlAlphabet}
\g@addto@macro{\UrlBreaks}{\UrlDigits}
\makeatother

\usepackage[pagebackref,breaklinks,colorlinks]{hyperref}

\usepackage[capitalize]{cleveref}
\crefname{section}{Sec.}{Secs.}
\Crefname{section}{Section}{Sections}
\Crefname{table}{Table}{Tables}
\crefname{table}{Tab.}{Tabs.}

\def\x{\times}

\renewcommand{\paragraph}[1]{\vspace{1.25mm}\noindent\textbf{#1}}

\newcommand\blfootnote[1]{%	
  \begingroup
  \renewcommand\thefootnote{}\footnote{#1}%
  \addtocounter{footnote}{-1}%
  \endgroup
}

\definecolor{sgreen}{RGB}{30, 150, 30} 
\definecolor{mygray}{gray}{0.92}
\definecolor{baselinecolor}{gray}{.9}
\newcommand{\baseline}[1]{\cellcolor{baselinecolor}{#1}}

\begin{document}
	
	%\title{Video Masked Autoencoders are Scalable and Transferrable Learners}
	\title{VideoMAE V2: Scaling Video Masked Autoencoders with Dual Masking}
	
	\author{
	   Limin Wang\textsuperscript{1,2,*} \quad Bingkun Huang\textsuperscript{1,2,*}  \quad Zhiyu Zhao\textsuperscript{1,2} \quad Zhan Tong\textsuperscript{1} \\ \quad Yinan He\textsuperscript{2}  \quad Yi Wang\textsuperscript{2}  \quad Yali Wang\textsuperscript{3,2}  \quad  Yu Qiao\textsuperscript{2,3} \\
		$^1$ State Key Laboratory for Novel Software Technology, Nanjing University, China \\
		$^2$ Shanghai AI Lab, China \quad
        $^3$ Shenzhen Institute of Advanced Technology, CAS, China \\
	}

	\maketitle
	\blfootnote{*~: Equal contribution.}
	%%%%%%%%% ABSTRACT
	\begin{abstract}
       Scale is the primary factor for building a powerful foundation model that could well generalize to a variety of downstream tasks. However, it is still challenging to train video foundation models with billions of parameters. 
       This paper shows that video masked autoencoder (VideoMAE) is a scalable and general self-supervised pre-trainer for building video foundation models. We scale the VideoMAE in both model and data with a core design. Specifically, we present a dual masking strategy for efficient pre-training, with an encoder operating on a subset of video tokens and a decoder processing another subset of video tokens. 
       Although VideoMAE is very efficient due to high masking ratio in encoder, masking decoder can still further reduce the overall computational cost. 
       This enables the efficient pre-training of billion-level models in video.
       We also use a progressive training paradigm that involves an initial pre-training on a diverse multi-sourced unlabeled dataset, followed by a post-pre-training on a mixed labeled dataset.
       Finally, we successfully train a video ViT model with a billion parameters, which achieves a new state-of-the-art performance on the datasets of Kinetics (90.0\% on K400 and 89.9\% on K600) and Something-Something (68.7\% on V1 and 77.0\% on V2). In addition, we extensively verify the pre-trained video ViT models on a variety of downstream tasks, demonstrating its effectiveness as a general video representation learner.

	\end{abstract}
	
	%%%%%%%%% BODY TEXT

	\section{Introduction}
	\label{sec:intro}
 
\begin{figure}[!t]
  \includegraphics[width=1\linewidth]{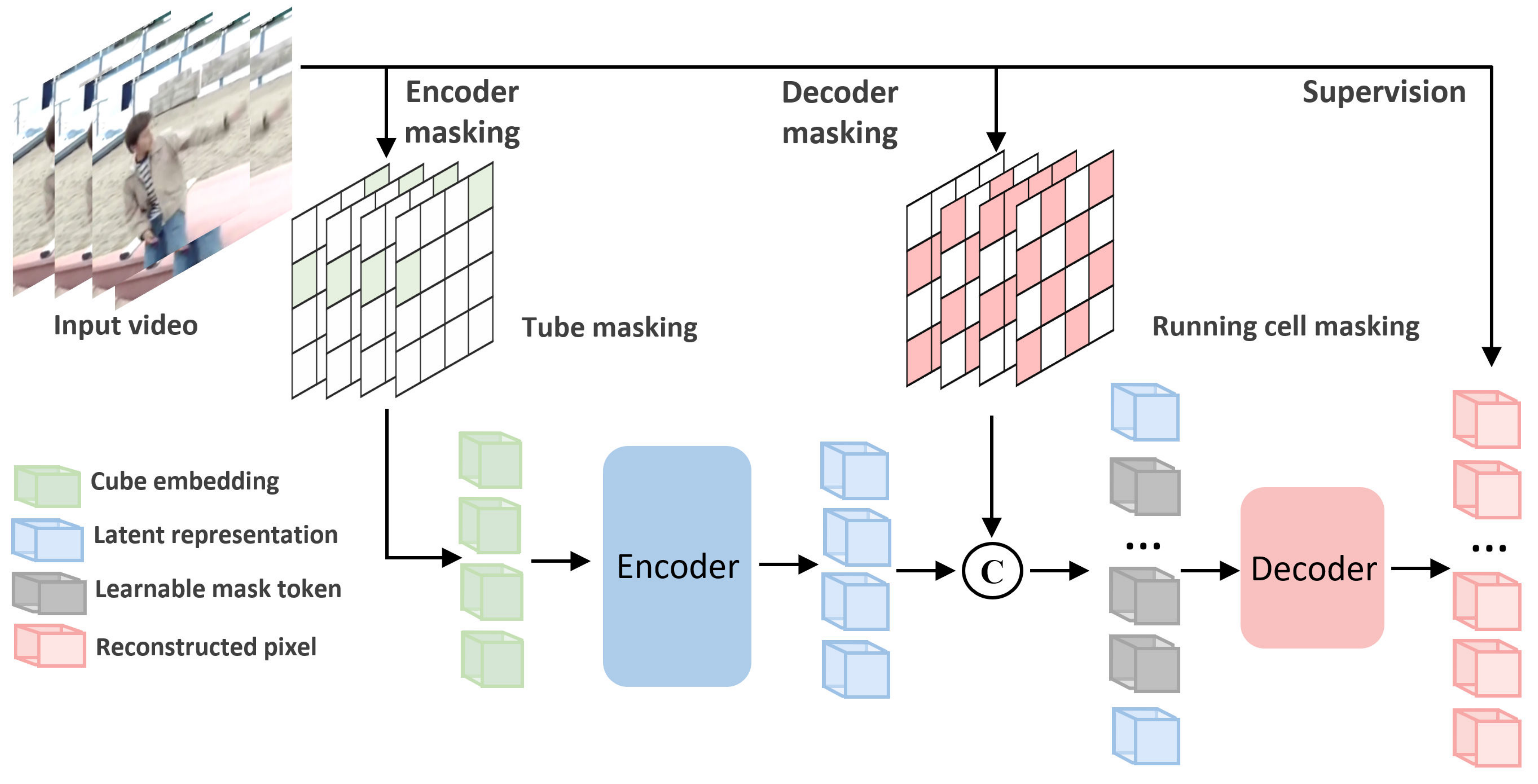}
  \vspace{-5mm}
  \caption{{\bf VideoMAE with dual masking.} To improve the overall efficiency of computation and memory in video masked autoencoding, we propose to mask the decoder as well and devise the dual masking strategy. Like encoder, we also apply a masking map to the deocoder and simply reconstruct a subset of pixel cubes selected by the running cell masking. The final reconstruction loss only applies for the invisible tokens dropped by the encoder.
}
  \label{fig:pipeline}
  \vspace{-5mm}
\end{figure}

    Effectively pre-training large foundation models~\cite{foundationmodel} on huge amounts of data is becoming a successful paradigm in learning generic representations for multiple data modalities (e.g., language~\cite{bert,gpt}, audio~\cite{abs-2204-12260,abs-2204-12768}, image~\cite{BEIT,simmim,imagemae}, video~\cite{maskedfeat,videomae,MAE-ST}, vision-language~\cite{clip,align}). These foundation models could be easily adapted to a wide range of downstream tasks through zero-shot recognition, linear probe, prompt tuning, or fine tuning. Compared with the specialized model to a single task, they exhibit excellent generalization capabilities and have become the main driving force for advancing many areas in AI.
	
    For vision research, many efforts have been devoted to developing effective pre-trained models. Among them, Transformer~\cite{transformer} with masked autoencoding~\cite{bert} is becoming a conceptually simple yet effective self-supervised visual learner (e.g., BEiT~\cite{BEIT}, SimMIM~\cite{simmim}, MAE~\cite{imagemae} for images, and MaskFeat~\cite{maskedfeat}, VideoMAE~\cite{videomae}, MAE-ST~\cite{MAE-ST} for videos). Meanwhile, based on the results in language models~\cite{gpt}, scaling model capacity and data size is an important ingredients for its remarkable performance improvement. However, for pre-trained vision models, very few work~\cite{swinv2} has tried to scale up this masked autoencoder pre-training to the billion-level models in image domain, partially due to the high data dimension and the high computational overhead. This issue is even more serious for scaling up video masked autoencoder pre-training owning to its extra time dimension and strong temporal variations.
	
    Following the promising findings in languages and images, we aim to {\em study the scaling property of video masked autoencoder (VideoMAE)}, and {\em push its performance limit on a variety of video downstream tasks}. We scale VideoMAE in both model and data. For model scaling, we try to instantiate the VideoMAE with vision transformer (ViT)~\cite{vit} having billion-level parameters (e.g., ViT-g~\cite{scalingvit}), and for data scaling, we hope to increase the pre-training dataset size to million-level to fully unleash the power of billion-level ViT model. However, to successfully train giant VideoMAE on such huge amounts of data and achieve impressive improvements on all considered downstream tasks, we still need to carefully address a few issues.
	
	First, we find computational cost and memory consumption is the bottleneck of scaling VideoMAE on the current GPUs with limited memory. Although VideoMAE~\cite{videomae} has improved its pre-training efficiency and reduced its memory consumption by employing the efficient asymmetric encoder-decoder architecture~\cite{imagemae} (i.e., dropping large numbers of tokens in encoder), it still fails to well support the billion-level video transformer pre-training. It takes more than two weeks to pre-train a ViT-g model with VideoMAE on 64 A100 GPUs. To further improve its pre-training efficiency, we find video data redundancy can be used to not only mask a high portion of cubes in the encoder, but also drop some cubes in the decoder. This solution yields higher pre-training efficiency and creates a similarly challenging and meaningful self-supervised task. In practice, it will increase the pre-training batchsize and reduce the pre-training time by a third with almost no performance drop.
	
	Second, MAE is still demanding for large data~\cite{datascaling} and billion-level video transformer tends to overfit on relatively small data. Unlike images, the existing public video dataset is much smaller. For example, there are only 0.24M videos in the Kinetics400 dataset~\cite{kinetics}, while the ImageNet-22k dataset~\cite{imagenet} has 14.2M images, let alone those publicly inaccessible image datasets such as JFT-3B~\cite{scalingvit}. Therefore, we need to come up with new ways to build a larger video pre-training dataset to well support the billion-level video transformer pre-training. We show that simply mixing the video datasets from multiple resources could produce an effective and diverse pre-training dataset for VideoMAE and improve its downstream performance of pre-trained models.
	
	Finally, it is still unknown how to adapt the billion-level pre-trained model by VideoMAE. Masked autoencoding is expected to learn invariant features that provide a favored initialization for vision transformer fine-tuning~\cite{umim}. However, directly fine-tuning billion-level pre-trained models on a relatively small video dataset (e.g., 0.24M videos) might be suboptimal, as the limited labeled samples might lead to overfitting issue in fine-tuning. In fact, in image domain, the intermediate fine-tuning technique~\cite{BEIT,swinv2} has been employed to boost the performance of masked pre-trained models.
    We show that collecting multiple labeled video datasets and building a supervised hybrid dataset can act as a bridge between the large-scale unsupervised dataset and the small-scale downstream target dataset. Progressive fine-tuning of the pre-trained models through this labeled hybrid dataset could contribute to higher performance in the downstream tasks.
	
	Based on the above analysis, we present a simple and efficient way to scale VideoMAE to billion-level ViT models on a dataset containing million-level pre-training videos. Our technical improvement is to introduce the {\em dual masking strategy} for masked autoencoder pipeline as shown in Figure~\ref{fig:pipeline}. In addition to the masking operation in encoder, we propose to mask decoder as well based on the data redundancy prior in video. With this dual-masked VideoMAE, we follow the intermediate fine-tuning in images~\cite{BEIT,swinv2}, and use a progressive training pipeline to perform the video masked pre-training on the million-level unlabeled video dataset and then post-pre-training on the labeled hybrid dataset. These core designs contribute to an efficient billion-level video autoencoding framework, termed as {\bf VideoMAE V2}. Within this framework, {\em we successfully train the first video transformer model with one billion parameters}, which attains a new state-of-the-art performance on a variety of downstream tasks, including action recognition~\cite{kinetics,sth,hmdb,ucf}, spatial action detection~\cite{ava,ava-kinetics}, and temporal action detection~\cite{thumos,fineaction}.
	
    \section{Related Work}
    \label{sec:related}
	
	\noindent {\bf Vision foundation models.} The term of foundation model was invented in~\cite{foundationmodel}. It refers to those powerful models that are pre-trained on broad data and can be adapted to a wide range of downstream tasks. Early research works in vision focused on pre-training CNNs~\cite{LeCunBBH98} or Transformers~\cite{transformer} on large-scale labeled datasets such as ImageNet-1k~\cite{KrizhevskySH12,resnet}, ImageNet-22k~\cite{swin,pvt}, and JFT~\cite{scalingvit}. Some recent works tried to perform unsupervised pre-training using contrastive learning~\cite{insdis,moco,SimCLR} or siamese learning~\cite{siamese}. Meanwhile, following the success in NLP~\cite{gpt,bert}, masked autoencoding was also introduced to pre-train image foundation models in a self-supervised manner, such as BEiT~\cite{BEIT}, SimMIM~\cite{simmim}, and MAE~\cite{imagemae}. Some vision-language pre-trained models, such as CLIP~\cite{clip} and ALIGN~\cite{align}, were proposed by learning from the alignment between images and text on noisy web-scale samples. They have shown excellent performance on zero-shot transfer.
	
	Concerning video foundation models, their progress lags behind images, partially due to the relatively smaller video datasets and higher complexity of video modeling. Since the introduction of Kinetics benchmarks~\cite{kinetics}, some supervised pre-trained models on it have been transferred to small-scale datasets for action recognition, such as 2D CNNs (TSN~\cite{tsn}, TSM~\cite{tsm}, TANet~\cite{tanet}, TDN~\cite{tdn}), 3D CNNs (I3D~\cite{i3d}, R(2+1)D~\cite{r2+1d}, ARTNet~\cite{artnet}, SlowFast~\cite{Slowfast}), Transformer (TimeSformer~\cite{timsformer}, Video Swin~\cite{videoswin}, UniFormer~\cite{uniformer}). Recently, some self-supervised video models are developed based on masked autoencoding such as BEVT~\cite{bevt}, MaskedFeat~\cite{maskedfeat}, VideoMAE~\cite{videomae}, and MAE-ST~\cite{MAE-ST} by directly extending these image masked modeling frameworks. However, these video foundation models often limit in their pre-training data size and model scale. More importantly, their downstream tasks have a narrow focus on action recognition, without consideration of other video tasks such as temporal action localization.
 
	\vspace{1mm}
	\noindent{\bf Masked visual modeling.} Early works treated masking in denoised autoencoders~\cite{VincentLBM08} or context inpainting~\cite{PathakKDDE16}. Inspired by the great success in NLP~\cite{bert,gpt}, iGPT~\cite{iGPT20} operated pixel sequences for prediction and ViT~\cite{vit} investigated the masked token prediction for self-supervised pre-training. Recently, there has been a surge of research into Transformer-based architectures for masked visual modeling~\cite{BEIT,imagemae,simmim,maskedfeat,bevt,videomae,MAE-ST}. BEiT~\cite{BEIT}, BEVT~\cite{bevt}, and VIMPAC~\cite{vimpac} learned visual representations by predicting discrete tokens. MAE~\cite{imagemae} and SimMIM~\cite{simmim} directly performed pixel masking and reconstruction for pre-training without discrete token representation. MaskFeat~\cite{maskedfeat} reconstructed the HOG~\cite{hog} features of masked tokens to perform self-supervised pre-training in videos. VideoMAE~\cite{videomae} and MAE-ST~\cite{MAE-ST} extended MAE~\cite{imagemae} to video domain for self-supervised video pre-training and achieved impressive performance on action recognition.

    \vspace{1mm}
	\noindent {\bf Vision model scaling.} Many works tried to scale up CNNs to improve recognition performance~\cite{inception,vggnet,resnet}. EfficientNet~\cite{TanL19} presented a scaling strategy to balance depth, width, and resolution for CNN design. Several works~\cite{MahajanGRHPLBM18,DBLP:conf/nips/HuangCBFCCLNLWC19,KolesnikovBZPYG20} tried to train much larger CNNs to obtain excellent performance by enlarging model capacities and training data size. Recently, a few works~\cite{scalingvit,swinv2} tried to scale up the vision transformer to the billion-level models with large-scale supervised pre-training on JFT-3B~\cite{scalingvit} or self-supervised pre-training on IN-22K-ext-70M~\cite{swinv2}. VideoMAE~\cite{videomae} and MAE-ST~\cite{MAE-ST} have trained the huge video transformer with millions of parameters. MAE-ST~\cite{MAE-ST} also tried the MAE pre-training on 1M IG-uncurated clips but failed to obtain better performance on Kinetics than small-scale pre-training. We are the first work to train video transformer with billion-level parameters.

    \section{VideoMAE V2}
	In this section, we first revisit VideoMAE and analyze its property. Then we present the dual masking strategy for the efficient training of VideoMAE. Finally, we present the scaling details of VideoMAE for large-scale pre-training. 
	
	\subsection{VideoMAE Revisited}
	\label{sec:videomae}
   We scale the video masked autoencoder (VideoMAE) due to its simplicity and high performance. 
	VideoMAE processes the downsampled frames $\mathbf{I} \in \mathbb{R}^{C \times T \times H \times W}$ from a clip with stride $\tau$, and uses the cube embedding $\Phi_{emb}$ to transform the frames into a sequence of tokens. Then, it designs a customized tube masking strategy to drop tokens with an extremely high ratio $\rho$ (e.g., 90\%). Finally, the unmasked tokens are fed into a video autoencoder $(\Phi_{enc}, \Phi_{dec})$ for reconstructing the masked pixels. 
        Specifically, VideoMAE is composed of {\em three core components}: cube embedding, encoder, and decoder. First, cube embedding encodes the local spatiotemporal features and builds the token list: $\mathbf{T} = \Phi_{emb} (\mathbf{I})$, where $\mathbf{T} = \{T_i\}_{i=1}^N$ is the token sequence, $T_i$ is the token produced by the embedding layer and then added with positional embedding, and $N$ is the total token number. Then the encoder simply operates on the {\em unmasked} tokens $\mathbf{T}^{u}$ with a vanilla ViT of joint space-time attention: $\mathbf{Z} = \Phi_{enc}(\mathbf{T}^{u})$, where $\mathbf{T}^u$ represents the unmasked visible tokens $\mathbf{T}^u = \{T_i\}_{i \in (1-\mathbb{M}(\rho))} $, $\mathbb{M}(\rho)$ is the masking map, and its token length $N^e$ is equal to $0.1N$. Finally, the decoder takes the {\em combined} tokens $\mathbf{Z}^c$ as inputs and performs reconstruction with another ViT: $\hat{\mathbf{I}} = \Phi_{dec}({\mathbf{Z}^c})$, where the combined tokens $\mathbf{Z}^c$ is the concatenated sequence of encoded token features $\mathbf{Z}$ and the learnable masked tokens [\texttt{MASK}] (with position embeddings), and its token length $N^d$ is equal to the original token number $N$. The loss function is the mean squared error (MSE) loss between the normalized masked pixels and the reconstructed pixels: $\ell = \frac{1}{\rho N}\sum_{i \in  \mathbb{M}(\rho)} |\mathbf{I}_i - \hat{\mathbf{I}}_i|^2$. 
	
	{\bf Computational cost analysis.} High efficiency is an important characteristic of masked autoencoder. 
	VideoMAE employs an asymmetric encoder-decoder architecture~\cite{imagemae}, where token sequence length of encoder is only one-tenth of decoder (i.e. $N^e=0.1N^d$). This smaller encoder input contributes to more efficient pre-training pipeline compared with other masked autoencoding frameworks~\cite{BEIT,simmim}. However, when scaling VideoMAE in both depth and width (channels) to a billion-level model, the overall computation and memory consumption is still the bottleneck for the current available GPUs with limited memory. Therefore, the current asymmetric encoder-decoder architecture needs to be further improved for scaling VideoMAE.

	\subsection{Dual Masking for VideoMAE}
	To better enable large-scale VideoMAE pre-training under a limited computational budget, we present a dual masking scheme to further improve its pre-training efficiency. As shown in Figure~\ref{fig:pipeline}, our dual masking scheme generates two masking maps $\mathbb{M}_e = \mathcal{M}_e(\rho^e)$ and $\mathbb{M}_d = \mathcal{M}_d(\rho^d)$ with two different masking generation strategies and masking ratios. These two masking maps $\mathbb{M}_e$ and $\mathbb{M}_d$ are for encoder and decoder, respectively. Like VideoMAE, our encoder operates on the partial and visible tokens under the encoder mask $\mathbb{M}_e$, and maps the observed tokens into latent feature representations. But unlike VideoMAE, our decoder takes inputs from the encoder visible tokens and {\em part of the remaining} tokens visible under the decoder mask $\mathbb{M}_d$. In this sense, we use the decoder mask to reduce the decoder input length for high efficiency yet attain similar information to the full reconstruction. Our decoder maps the latent features and the remaining incomplete tokens into the pixel values at the corresponding locations. The supervision only applies to the decoder output tokens invisible to the encoder. We will detail the design next. 
	
	\paragraph{\bf Masking decoder.} As analyzed in Section~\ref{sec:videomae}, the decoder of VideoMAE is still inefficient as it needs to process all the cubes in videos. Thus, we further explore the prior of data redundancy in the decoder and propose the strategy of {\em masking decoder}. Our idea is mainly inspired by the recent efficient action recognition transformer~\cite{MAR}, which only uses a small portion of tokens to achieve similar performance. It implies data redundancy exists in inference, which applies for our reconstruction target as well.
	
	Our dual masking strategy is composed of encoder masking $\mathcal{M}_e$ and decoder masking $\mathcal{M}_d$. The encoder masking is the random tube masking with an extremely high ratio, which is the same as the original VideoMAE. For decoder masking, our objective is opposite to encoder masking. The tube masking in encoder tries to relieve the issue of ``information leakage'' caused by temporal correlation. In contrast, in decoder masking, we need to encourage  ``information complement'' to ensure minimal information loss in this partial reconstruction. In this sense, we need to select as diverse cubes as possible to cover the whole video information. In the implementation, we compare different masking strategies and eventually choose the running cell masking~\cite{MAR}. With this decoder masking map $\mathbb{M}_d$~\footnote{For a clear presentation, unlike encoder, we use this masking map to denote the kept and visible tokens in decoder input.}, we reduce the decoder input length to improve efficiency.

	\paragraph{\bf VideoMAE with dual masking.} Our improved VideoMAE shares the same cube embedding and encoder with the original VideoMAE as described in Section~\ref{sec:videomae}. For decoder, it processes the combined tokens of encoder output and part the remaining visible tokens under the decoder mask $\mathbb{M}_d$. Specifically, the combined sequence is defined as:
	\begin{equation}
	    \mathbf{Z}^c = \mathbf{Z} \cup \{\mathbf{M}_i\}_{i \in \mathbb{M}_d},
	\end{equation}
	where $\mathbf{Z}$ is the latent representation from encoder, $\mathbf{M}_i$ is the learnable masking token with corresponding positional embedding. With this combined token sequence $\mathbf{Z}^c$, our decoder only reconstructs the visible tokens under the decoder mask. The final MSE loss is computed between the normalized masked pixels $\mathbf{I}$ and the reconstructed ones $\hat{\mathbf{I}}$ over the decoder visible cubes:
	\begin{equation}
	    \ell = \frac{1}{(1-\rho^d) N} \sum_{i \in \mathbb{M}_d \cap \mathbb{M}_e} |\mathbf{I}_i - \hat{\mathbf{I}}_i|^2.
	\end{equation}
	
	\subsection{Scaling VideoMAE}
	
	\paragraph{\bf Model scaling.} Model scale is the primary force in obtaining excellent performance. Following the original VideoMAE, we use the vanilla ViT~\cite{vit} as the backbone due to its simplicity. According to the scaling law of ViT~\cite{scalingvit}, we build VideoMAE encoder with backbones of different capacities ranging from ViT-B, ViT-L, ViT-H, to ViT-g. Note that {\em ViT-g is a large model with billion-level parameters and has never been explored in video domain}. Its performance with masked autoencoding for video representation learning is still unknown to the community. More details on these backbone designs could be referred to~\cite{scalingvit}. For decoder design, we use relatively shallow and lightweight backbones~\cite{videomae,imagemae} with fewer layers and channels. In addition, we apply our dual masking strategy to further reduce computational cost and memory consumption. More details on the decoder design could be found in the appendix.
	
	\paragraph{\bf Data scaling.} Data scale is another important factor that influences the performance of VideoMAE pre-training. The original VideoMAE simply pre-train the ViT models on relatively small-scale datasets by emphasizing its data efficiency. In addition, they require to pre-train the individual models specific to each dataset (i.e., Something-Something and Kinetics datasets have different pre-trained models). In contrast, we aim to learn a universal pre-trained model that could be transferred to different downstream tasks. To this end, we try to increase the pre-training video samples to a million-level size and aim to understand the data scaling property for VideoMAE pre-training. Data diversity is important for learning general video representations. Therefore, we build an {\em unlabeled hybrid} video dataset covering videos from General Webs, Youtube, Instagram, Movies, and Manual Recordings. We collect videos from the public datasets of Kinetics, Something-Something, AVA, WebVid, and uncurated videos crawled from Instagram. In total, there are 1.35M clips in our unlabeled mixed dataset. Note that {\em pre-training video transformer on a such large-scale and diverse dataset is rare in previous works} and {\em it still remains unknown the influence of data scale and diversity on VideoMAE pre-training.} More details on our dataset could be found in the appendix.
	
	\paragraph{\bf Progressive training.} Transferring scheme is an important step to adapt the pre-trained large video transformers to the downstream tasks. The masked autoencoder pre-training is expected to learn some invariant features and can provide a favored initialization for vision transformer fine-tuning~\cite{umim}. The original VideoMAE directly fine-tunes the pre-trained models on the target dataset only with its corresponding supervision. This direct adapting strategy might fail to fully unleash the power of large pre-trained video transformer due to limited supervision. Instead, in order to relieve the overfitting risk, we argue that {\em we should leverage the semantic supervision signals from multiple sources in multiple stages to gradually adapt the pre-trained video transformers to downstream tasks}. Accordingly, following the intermediate fine-tuning in images~\cite{BEIT,swinv2}, we devise a {\em progressive training} pipeline for the whole training process of billion-level video transformers. First, we conduct unsupervised pre-training with masked autoencoding on the unlabeled hybrid video dataset. Then, we build a {\em labeled hybrid} dataset by collecting and aligning multiple existing supervised datasets with labels. We perform the supervised post-pre-training stage on this labeled hybrid dataset to incorporate the semantics from multiple sources into the previous pre-trained video transformers. Finally, we perform the {\em specific fine-tuning} stage on the target dataset to transfer the general semantics to the task-centric knowledge.
	
	Based on the above designs of dual masking, data scaling, and progressive training, we implement a simple and efficient masked autoencoding framework with a billion-level ViT backbone, termed as {\em VideoMAE V2}. With this new framework, we successfully train the first billion-level video transformer and push the vanilla ViT performance limit on a variety of video downstream tasks, including video action recognition, action detection, and temporal action detection.

	\section{Experiments}

\subsection{Implementation and Downstream Tasks}
\paragraph{\bf Model.} We conduct investigations on the VideoMAE V2 by scaling its model capacity and pre-training data size. We scale the backbone network from the existing huge ViT model (ViT-H) to the giant ViT model (ViT-g)~\cite{scalingvit}. The ViT-g has a smaller patch size (14), more encoder blocks (40), a higher dimension of cube embedding and self-attention (1408), and more attention heads (16). It has 1,011M parameters. More details could be referred to~\cite{scalingvit}. 

\paragraph{\bf Data.} To well support the billion-level ViT model pre-training, we build two large-scale video datasets for our proposed progressive training. For self-supervised pre-training of VideoMAE V2, we build a million-level unlabeled video dataset by collecting clips from multiple resources such as Movie, Youtube, Instagram, General Webs, and manual recordings from scripts, and the dataset is termed as {\em UnlabeledHybrid}. Specifically, our dataset is built by simply selecting videos from the public available datasets of Kinetics~\cite{kinetics}, Something-Something~\cite{sth}, AVA~\cite{ava}, WebVid2M~\cite{webvid2m}, and our own crawled Instagram dataset. In total, there are around 1.35M clips in our mixed dataset and this is the largest dataset ever used for video masked autoencoding. For supervised post-pre-training, we collect the larger video dataset with human annotations, termed as {\em LabeledHybrid}. Following~\cite{uniformerv2}, we take the union of different versions of Kinetics datasets (K400, K600, K700) by aligning their label semantics and removing the duplicate videos with the validation sets. This labeled hybrid dataset has 710 categories and 0.66M clips. We pre-train our video transformer model on these two datasets and then transfer them to the downstream tasks as detailed next. More details on these pre-training datasets could be found in the appendix.

\paragraph{\bf Tasks.} To verify the generalization ability of VideoMAE V2 pre-trained ViTs as video foundation models, we transfer their representations to a variety of downstream tasks. 

{\em Video Action Classification.} Action classification is the most common task in video understanding. Its objective is to classify each trimmed clip into a predefined action class and evaluated the average accuracy over action classes. According to the original VideoMAE~\cite{videomae}, we perform detailed analysis on this task to investigate the property of scaling video masked autoencoding. In experiments, we choose four datasets to report its performance: Kinetics~\cite{kinetics}, Something-Something~\cite{sth}, UCF101~\cite{ucf}, and HMDB51~\cite{hmdb}. Kinetics and Something-Something are two large-scale action recognition datasets and have their own unique property for action recognition, where Kinetics contains appearance-centric action classes while Something-Something focuses on motion-centric action understanding. UCF101 and HMDB51 are two relatively small datasets and suitable to verify the transfer performance of large pre-trained models as shown in the appendix.

{\em Spatial Action Detection.} Action detection is an important task in video understanding, and it aims to recognize all action instances and localize them in space. This task is more challenging than action classification as it deals with more fine-grained action classes and needs to capture detailed structure information to discriminate co-occurring action classes. In experiments, we choose two action detection benchmarks to illustrate the effectiveness of our pre-trained models by VideoMAE V2, namely AVA~\cite{ava} and AVA-Kinetics~\cite{ava-kinetics}. AVA contains the box annotations and their corresponding action labels on keyframes (could be more than one label for each human box). The annotations are done at 1FPS over 80 atomic classes. AVA-Kinetics introduces the AVA style annotations to the Kinetics dataset and a single frame of selected video from Kinetics is annotated with AVA labels. The evaluation metric is frame-level Average  Precision (mAP) under the IoU threshold of 0.5.

{\em Temporal Action Detection.} Temporal action detection is an important task in long-form video understanding. Its goal is to recognize all action instances in an untrimmed video and localize their temporal extent (starting and ending timestamps). Unlike spatial action detection, temporal action localization aims to focus on precise temporal boundary localization. Intuitively, in addition to capturing semantic information for recognition, the pre-trained models should be able to effectively model the temporal evolution of features to detect action boundaries. In experiments, we choose two temporal action detection benchmarks to evaluate the performance of our pre-trained video models: THUMOS14~\cite{fineaction} and FineAction~\cite{fineaction}. THUMOS14 is a relatively small and well labeled temporal action detection dataset, that has been widely used by the previous methods. It only includes sports action classes on this dataset. FineAction is a new large-scale temporal action dataset with fine-grained action class definitions. The evaluation metric is the average mAP under different tIoU thresholds.

\subsection{Main Results}
\begin{table}[t]
    \centering
    \small
    \setlength{\tabcolsep}{10pt}
    \renewcommand\arraystretch{1.0}
    \begin{tabular}{cc|cc}
        Decoder Masking & $\rho^d$ &  Top-1 & FLOPs\\
        \shline
        None & 0\% & {\bf 70.28} & 35.48G \\
        Frame & 50\% & 69.76 & 25.87G \\ 
        Random & 50\% & 64.87 & 25.87G \\
        Running cell~\textcolor{red}{\footnotesize{$^1$}} & 50\% & 66.74 & 25.87G \\
        \shline
        Running cell~\textcolor{red}{\footnotesize{$^2$}} & 25\% & 70.22 & 31.63G \\
        \baseline{Running cell~\textcolor{red}{\footnotesize{$^2$}}} & \baseline{50\%} & \baseline{70.15} & \baseline{25.87G} \\
        Running cell~\textcolor{red}{\footnotesize{$^2$}} & 75\% & 70.01 & 21.06G \\
    \end{tabular}
    \vspace{-0.8em}
    \captionof{table}{{\bf Ablation study on the decoder masking strategies.} 
    Experiments are conducted with ViT-B by pre-training on SSv2 with 800 epochs. ``None'' refers to the original VideoMAE without decoder masking. We use a better fine-tuning setting than the original VideoMAE.
    \textcolor{red}{\footnotesize{$^1$}} Loss computed over all decoder output tokens.
    \textcolor{red}{\footnotesize{$^2$}} Loss computed over only decoder output tokens invisible to encoder. The default setting for VideoMAE v2 is colored in \colorbox{baselinecolor}{gray}.
    }\label{tab:ablation-deooder-masking}
    \vspace{-5mm}
\end{table}

    \begin{table*}[h!]
    \centering
    \footnotesize
    \setlength{\tabcolsep}{10pt}
    \renewcommand\arraystretch{1.0}
    \begin{tabular}{c|ccccccc}
        Masking & Backbone & pre-training dataset & FLOPs & Mems & Time &  Speedup & Top-1\\
        \shline
        Encoder masking & ViT-B & Something-Something V2 & 35.48G & 631M & 28.4h & - & 70.28\\
        Dual masking & ViT-B & Something-Something V2 & 25.87G & 328M &  15.9h &  {\bf 1.79}$\times$ &70.15 \\ 
        % \hline
        Encoder masking & ViT-g & UnlabeledHybrid & 263.93G & 1753M & 356h\textcolor{red}{\footnotesize{$^1$}} & - & -\\
        Dual masking & ViT-g & UnlabeledHybrid & 241.61G & 1050M & 241h & {\bf 1.48}$\times$ & 77.00\\ 
    \end{tabular}
    \vspace{-1.2em}
    \captionof{table}{{\bf Comparison between dual masking and encoder-only masking}. We report the computational cost, memory consumption, and running time for comparison. We perform experiments with backbones (ViT-B and ViT-g) and pre-training on two scales of datasets (Sth-Sth V2 and UnlabeledHybrid). Time is for 1200 epochs on 64 GPUs. \textcolor{red}{\footnotesize{$^1$}} is estimated by training 5 epochs.
    }\label{tab:ablation-dual-masking}
    \vspace{-3mm}
    \end{table*}

\begin{table*}[h!]
    \centering
    \footnotesize
    \setlength{\tabcolsep}{8pt}
    \renewcommand\arraystretch{1.0}
    \begin{tabular}{c|cccccccc}
        method & pre-train data & data size & epoch & ViT-B & ViT-L & ViT-H & ViT-g \\
        \shline
        MAE-ST~\cite{MAE-ST}  & Kinetics400 & 0.24M & 1600 & 81.3 & 84.8 & 85.1 & - \\
        MAE-ST~\cite{MAE-ST} & IG-uncurated & 1M & 1600 & - & 84.4 & - & -  \\
        VideoMAE V1~\cite{videomae} & Kinetics400 & 0.24M  & 1600 & {\bf 81.5} & 85.2 & 86.6 & - \\
        VideoMAE V2 & UnlabeledHybrid & 1.35M &  1200 & {\bf 81.5} (77.0)  & {\bf 85.4} (81.3) & {\bf 86.9} (83.2) & {\bf 87.2} (83.9)  \\
        $\Delta$\textcolor{sgreen}{$Acc.$} with V1 & -& - & - & \textcolor{sgreen}{\textit{\textbf{+ {0}}\%}} & \textcolor{sgreen}{\textit{\textbf{+ 0.2}\%}} & \textcolor{sgreen}{\textit{\textbf{+ 0.3}\%}} & - \\

    \end{tabular}
    \vspace{-3mm}
    \captionof{table}{{\bf Results on the Kinetics-400 dataset}. We scale the pre-training of VideoMAE V2 to billion-level ViT-g model with million-level data size. We report the fine-tuning accuracy of multiple view fusion (5$\times$3) and single view results in the bracket. All models are pre-trained and fine-tuned at the input of 16$\times$224$\times$224 and sampling stride $\tau=4$.
    }\label{tab:kinetics-result}
    \vspace{-3mm}
    \end{table*}

    \begin{table*}[h!]
    \centering
    \footnotesize
    \setlength{\tabcolsep}{8pt}
    \renewcommand\arraystretch{1.0}
    \begin{tabular}{c|cccccccc}
        method & pre-train data & data size & epoch & ViT-B & ViT-L & ViT-H & ViT-g \\
        \shline
        MAE-ST~\cite{MAE-ST}  & Kinetics400 & 0.24M & 1600 & - & 72.1 & 74.1 & -  \\
        MAE-ST~\cite{MAE-ST}  & Kinetics700 & 0.55M & 1600 & - & 73.6 & 75.5 & -  \\
        VideoMAE V1~\cite{videomae} & Something-Something V2 & 0.17M  & 2400 & 70.8 & 74.3 &  74.8 & -  \\
        VideoMAE V2 & UnlabeledHybrid & 1.35M &  1200 & {\bf 71.2} (69.5) & {\bf 75.7} (74.00) & {\bf 76.8} (75.5) & {\bf 77.0} (75.7)  \\
        $\Delta$\textcolor{sgreen}{$Acc.$} with V1 & - &- & - & \textcolor{sgreen}{\textit{\textbf{+ {0.4}}\%}} & \textcolor{sgreen}{\textit{\textbf{+ 1.4}\%}} & \textcolor{sgreen}{\textit{\textbf{+ 2.0}\%}} & - \\

    \end{tabular}
    \vspace{-3mm}
    \captionof{table}{{\bf Results on the Something-Something V2 dataset}. We scale the pre-training of VideoMAE V2 to billion-level ViT-g model with million-level data size. We report the fine-tuning accuracy of multiple view fusion (2$\times$3) and single view results in the brackets. All models are pre-trained at input of 16$\times$224$\times$224 and sampling stride $\tau=2$. Fine-tuning is on the same size as TSN~\cite{tsn} sampling.
    }\label{tab:sth-result}
    \vspace{-2mm}
    \end{table*}

  \begin{table}[h!]
    \centering
    \footnotesize
    \setlength{\tabcolsep}{5pt}
    \renewcommand\arraystretch{1.0}
    \begin{tabular}{c|ccc}
        method & extra supervision  & ViT-H & ViT-g   \\
        \shline
        MAE-ST~\cite{MAE-ST}  & K600  & 86.8  & - \\
        VideoMAE V1~\cite{videomae} & K710 & 88.1 (84.6) & - \\
        % \hline
        VideoMAE V2 & - &  86.9 (83.2) & 87.2 (83.9)  \\
        VideoMAE V2 & K710 &   {\bf 88.6}  (85.0) & {\bf 88.5} (85.6) \\
        $\Delta$\textcolor{sgreen}{$Acc.$} with V1 & K710& \textcolor{sgreen}{\textit{\textbf{+ 0.5}\%}} & - \\

    \end{tabular}
    % \vspace{-1em}
    \captionof{table}{{\bf Study on progressive pre-training}. We report the fine-tuning accuracy of multiple view fusion (5$\times$3) and single view results in the bracket on the Kinetics-400 dataset. The implementation detail is the same with Table~\ref{tab:kinetics-result}.
    }\label{tab:progressive-training}
    \vspace{-7mm}
    \end{table}

    \begin{table*}[t]
	\begin{subtable}[t]{.39\linewidth}
		\centering
		\caption{Kinetics 400}
        \setlength{\tabcolsep}{3pt} %
		\renewcommand*{\arraystretch}{1.04}  %

		\scriptsize{
			\begin{tabular}{lcccc}
				\toprule
				Method & Top 1 & Top 5         & Views 	& TFLOPs \\
				\midrule
				I3D NL~\cite{nonlocal}										 & 77.7                  & 93.3         		 &  $10 \times 3$ & 10.77     \\
                TDN~\cite{tdn} & 79.4 & 94.4 &  $10 \times 3$ &  5.94 \\
				SlowFast R101-NL~\cite{Slowfast}       		&  79.8                 &  93.9                   & $10 \times 3$    & 7.02   \\  %
				TimeSformer-L~\cite{timsformer} & 80.7 & 94.7 & $1 \times 3$ & 7.14 \\
                MTV-B ($320^2$)~\cite{mtv} & 82.4 & 95.2 	& $4 \times 3$ & 11.16 \\
                Video Swin-L ($384^2$)~\cite{videoswin} & 84.9 & 96.7 & $10 \times 5$ & 105.35 \\
				ViViT-L FE~\cite{vivit} & 81.7 	&  93.8  & $1 \times 3$ &  11.94 \\  %
                MViTv2-L ($312^2$)~\cite{mvitv2} & 86.1 & 97.0 & $40 \times 3$ & 42.42 \\
                MaskFeat~\cite{maskedfeat} & 87.0 & 97.4 & $4 \times 3$ & 45.48 \\
                MAE-ST~\cite{MAE-ST} & 86.8 & 97.2 &  $4 \times 3$ &  25.05 \\
                VideoMAE~\cite{videomae} & 86.6 & 97.1  & $5 \times 3$ & 17.88 \\
                {\bf VideoMAE V2-H} & 88.6 & 97.9 & $5 \times 3$ & 17.88 \\
                {\bf VideoMAE V2-g} & 88.5 & 98.1 & $5 \times 3$ & 38.16 \\
                {\bf VideoMAE V2-g ($64\x266^2$)} & {\bf 90.0} & {\bf 98.4} & $2 \times 3$ & 160.30 \\
				\midrule
				\multicolumn{4}{l}{\textit{Methods using in-house labeled data}} \\ 
				\textcolor{gray}{CoVeR (JFT-3B)~\cite{zhang2021}} & \textcolor{gray}{87.2} & \textcolor{gray}{--} & \textcolor{gray}{$1 \times 3$} & \textcolor{gray}{-} \\
				\textcolor{gray}{\textbf{MTV-H} (WTS $280^2$)~\cite{mtv}} &  \textcolor{gray}{89.9} &  \textcolor{gray}{98.3} 	& \textcolor{gray}{$4 \times 3$} & \textcolor{gray}{73.57} \\ %
				\bottomrule
			\end{tabular}
		}
		\label{tab:sota_kinetics400}
		\centering
		\caption{Kinetics 600}
		\vspace{0.2\baselineskip}
		\setlength{\tabcolsep}{3pt} %
		\scriptsize{
			\begin{tabular}{lcccc}
				\toprule
				Method 																			 & Top 1                & Top 5         & Views 	& TFLOPs \\
				\midrule
				
				SlowFast R101-NL~\cite{Slowfast}       		&  81.8                &  95.1                  & $10 \times 3$    & 7.02   \\  %
				TimeSformer-L~\cite{timsformer} & 82.2 & 95.6 & $1 \times 3$ & 7.14 \\
                MTV-B ($320^2$)~\cite{mtv} & 84.0 & 96.2 & $4 \times 3$ & 11.16 \\
                ViViT-L FE~\cite{vivit} & 82.9 	&  94.6  & $1 \times 3$ &  11.94 \\  %
				MViTv2-L ($352^2$)~\cite{mvitv2} & 87.9 & 97.9 & $40 \times 3$ & 45.48 \\
                MaskFeat~\cite{maskedfeat} & 86.4 & 97.4 & $1 \times 10$ & 3.77 \\
                {\bf VideoMAE V2-H} & 88.3 & 98.1 & $5 \times 3$ & 17.88 \\
                {\bf VideoMAE V2-g} & 88.8 & 98.2 & $5 \times 3$ & 38.16 \\
                {\bf VideoMAE V2-g ($64\x266^2$)} & {\bf 89.9} & {\bf 98.5} & $2 \times 3$ & 160.30 \\
                
				\midrule
				\multicolumn{4}{l}{\textit{Methods using in-house labeled data}}                                \\ 
				\textcolor{gray}{CoVeR (JFT-3B)~\cite{zhang2021}} & \textcolor{gray}{87.9} & \textcolor{gray}{97.8} & \textcolor{gray}{$1 \times 3$} & \textcolor{gray}{-}\\
				\textcolor{gray}{\textbf{MTV-H} (WTS $280^2$)} & \textcolor{gray}{\textbf{90.3}} & \textcolor{gray}{\textbf{98.5}} & \textcolor{gray}{$4 \times 3$} & \textcolor{gray}{73.57} \\ %
				\bottomrule
			\end{tabular}
			\label{tab:sota_K600}
		}
	\end{subtable}
 \hspace{1.2mm}
  	%\hfill
  	\begin{subtable}[t]{.295\linewidth}
		\centering
  		\caption{Something-Something V2}
  		\setlength{\tabcolsep}{9pt} %
  		\scriptsize{
	  		\begin{tabular}{lcc}
	  			\toprule
	  			Method & Top 1 & Top 5      \\ %
	  			\midrule
	  			SlowFast~\cite{Slowfast}       		& 63.1                    &  87.6  \\ %
                    TEINet~\cite{teinet} & 66.5 & - \\
                    TEA~\cite{tea} & 65.1 & 89.9 \\
                    TDN~\cite{tdn} & 69.6 & 92.2 \\ 
	  			TimeSformer-L~\cite{timsformer}				  & 62.4				& -		\\ %
	  			MFormer-HR~\cite{motionformer} & 68.1 & 91.2  \\
	  			ViViT-L FE~\cite{vivit} & 65.9 	&  89.9 \\  %
                    Video Swin-B~\cite{videoswin} & 69.6 & 92.7 \\
	  			MViTv2-B~\cite{mvitv2}					  & 72.1				& 93.4		\\ %
	  			MTV-B~\cite{mtv} & 67.6 & 90.1 \\ %
                    BEVT~\cite{bevt} & 70.6 & - \\
                    VIMPAC~\cite{vimpac} & 68.1 & - \\
                    UniFormer~\cite{uniformer} & 71.2 & 92.8 \\
                    MaskFeat~\cite{maskedfeat} & 75.0 & 95.0 \\
                    MAE-ST~\cite{MAE-ST} & 75.5 & 95.0 \\
                    VideoMAE~\cite{videomae} & 75.4 & 95.2 \\
                    \midrule
                    {\bf VideoMAE V2-H} & 76.8 & 95.8 \\
                    {\bf VideoMAE V2-g} & {\bf 77.0} &  {\bf 95.9} \\
	  			\bottomrule
	  		\end{tabular}
	  		\label{tab:sota_sthsthv2}
  		}
		\centering
		\caption{Something-Something V1}
		\vspace{-0.1\baselineskip}
		\setlength{\tabcolsep}{9pt} %
		\scriptsize{
			\begin{tabular}{lcc}
				\toprule
				Method 												  & Top 1                & Top 5   \\
				\midrule
                I3D~\cite{i3d} & 41.6 & 72.2 \\
                NL I3D+GCN~\cite{WangG18} & 46.1 & 76.8 \\
                TSM~\cite{tsm} & 49.7 & 78.5 \\
                V4D~\cite{v4d} & 50.4 & - \\
                TANet~\cite{tanet} & 50.6 & 79.3 \\
                TEINet~\cite{teinet} & 52.5 & - \\
                TEA~\cite{tea} & 51.9 & 80.3 \\
                CorrNet~\cite{corrnet} & 53.3 & - \\
                GSM~\cite{GSM} & 55.2 & - \\
                TDN~\cite{tdn} & 56.8 & 84.1\\
                UniFormer~\cite{uniformer} & 61.0 & 87.6 \\
                \midrule
                {\bf VideoMAE V2-H} & 66.6 & 90.8 \\
                {\bf VideoMAE V2-g} & {\bf 68.7} & {\bf 91.9}   \\

				\bottomrule
			\end{tabular}
			\label{tab:sota_ssv1}
		}
  	\end{subtable}
  	\begin{subtable}[t]{.295\linewidth}
  		\setlength{\tabcolsep}{4pt} %
  		\centering
  		\caption{AVA}
  		\setlength{\tabcolsep}{6pt} %
  		\scriptsize{
  			\begin{tabular}{lcc}
  				\toprule
  				Method & Long Feature & mAP \\ 
  				\midrule
  				SlowFast~\cite{Slowfast} & \xmark & 29.0 \\
                    TubeR~\cite{tuber} & \cmark & 33.4 \\
  				MaskFeat~\cite{maskedfeat} & \xmark & 38.8 \\
                    MAE-ST~\cite{MAE-ST} & \xmark & 39.0 \\
  				VideoMAE~\cite{videomae}	&  \xmark    &  39.5    \\
  				\textbf{VideoMAE V2} & \xmark   & {\bf 42.6} \\  %
  				\bottomrule
  			\end{tabular}
  			\label{tab:sota-ava}
  		}
    
  		\vspace{0.18\baselineskip} %
  		\centering
  		\caption{AVA Kinetics}
  		\setlength{\tabcolsep}{6pt} %
  		\scriptsize{
  			\begin{tabular}{lcc}
  				\toprule
  				Method & Ensembled & mAP \\ 
  				\midrule
  				AIA++~\cite{xia2020report} & \cmark & 29.0 \\
                    MSF~\cite{Zhu2020report} & \cmark & 33.4 \\
  				ACAR~\cite{ACAR} & \cmark & 40.5 \\
  				{\bf VideoMAE V2}	&  \xmark    &  {\bf 43.9}    \\
  				\bottomrule
  			\end{tabular}
  			\label{tab:sota-ava-kinetics}
  		}
		\label{tab:sota_ava_kinetics}
  
  		\centering
  		\caption{THUMOS14}
  		\setlength{\tabcolsep}{6pt} %
  		\scriptsize{
  			\begin{tabular}{lcc}
  				\toprule
                  Method & Optical Flow & mAP \\ 
                 \midrule
                    RTD-Net~\cite{rtd} & \cmark & 43.6 \\
                    DaoTAD~\cite{rgb_enough} & \xmark & 50.0 \\
                    AFSD~\cite{afsd} & \cmark & 52.0 \\
  				DCAN~\cite{dcan} & \cmark & 52.3 \\
                    TadTR~\cite{TADTR} & \cmark & 54.2 \\
                    TALLFormer~\cite{tallformer} & \xmark & 59.2 \\
                    BasicTAD~\cite{basictad} & \xmark & 59.6 \\
  			    ActionFormer~\cite{actionformer} & \cmark & 66.8 \\
                   {\bf VideoMAE V2}	&  \xmark    &  {\bf 69.6} \\
  				\bottomrule
  			\end{tabular}
  			\label{tab:sota_thumos}
  		}

      		\centering
  		\caption{FineAction}
  		\setlength{\tabcolsep}{6pt} %
  		\scriptsize{
  			\begin{tabular}{lcc}
  				\toprule
      Method & Optical Flow & mAP \\ 
      \midrule
                    BMN~\cite{bmn} & \cmark & 9.25 \\
                    G-TAD~\cite{g-tad} & \cmark & 9.06 \\
  				BasicTAD~\cite{basictad} & \xmark & 12.2 \\
                    ActionFormer~\cite{actionformer} & \xmark & 13.2 \\
                   {\bf VideoMAE V2}	&  \xmark    &  {\bf 18.2} \\
  				\bottomrule
  			\end{tabular}
  			\label{tab:sota_fineaction}
  		}

	\end{subtable}
 \vspace{-1mm}
  \caption{{\bf Systematic study on the transfer performance of VideoMAE V2 pre-trained models.} We use them as the video foundation models and transfer them to three kinds of downstream tasks: action classification, action detection, and temporal action detection, covering eight mainstream video action benchmarks. Entries using extra in-house labeled data for training are in gray. ``-'': numbers not available.}
	\label{tab:sota}
	\vspace{-\baselineskip}
\end{table*}

We first conduct the experimental study on the core designs of our VideoMAE V2 pre-training framework. We report the fine-tuning action recognition accuracy on the datasets of Kinetics-400 and Something-Something (Sth-Sth) V2. Implementation details about the pre-training and fine-tuning could be found in the appendix.

\paragraph{Results on dual masking.} 
We first perform an ablation study on the decoder masking strategy. In this study, we use the ViT-B as the backbone and the pre-training is performed on the Sth-Sth V2 dataset with 800 epochs. The results are evaluated with the fine-tuning accuracy on the Sth-Sth V2 and reported in Table~\ref{tab:ablation-deooder-masking}. We first re-implement the original VideoMAE without decoder masking and achieve slightly better performance (70.28\% vs. 69.6\% Top-1 acc.). Then, we try two masking alternatives in decoder: frame masking and random masking. For frame masking, we only reconstruct half of the frames in the decoder, and for random masking, we stochastically drop half of the cubes in the decoder for reconstruction. These two alternatives perform worse than the original VideoMAE. Finally, we apply the running cell masking to select a subset of representative cubes for reconstruction. In this setting, we apply loss functions computed over all decoder output tokens or only encoder-invisible tokens. Agreed with the results in MAE and VideoMAE, the loss on all tokens performs worse partially due to information leakage of these visible tokens from encoder. The running cell masking scheme performs on par with the original result (70.28\% vs. 70.15\% top-1 acc.). We also ablates the decoder masking ratio and $50\%$ keeps a good trade-off between accuracy and efficiency.

In Table~\ref{tab:ablation-dual-masking}, we report the computational cost (FLOPs), memory consumption (Mems), and per-epoch running time (Time) of dual masking, and compare with the original encoder-only masking in VideoMAE. In this comparison, we use a lightweight decoder only with 4 transformer blocks (channel 384 for ViT-B and 512 for ViT-g). Our dual masking can further improve the computational efficiency of the original asymmetric encoder-decoder architecture in MAE~\cite{imagemae}. For memory consumption, we can reduce almost half of the overall memory of feature maps, and this is particularly important for pre-training billion-level video transformer under the available GPUs of limited memory. 

\paragraph{Results on data scaling.} We study the influence of pre-training data size on the VideoMAE V2 pre-training. In this experiment, we pre-train the video models with backbones from ViT-B, ViT-L, ViT-H, to ViT-g on our built UnlabeledHybrid dataset with around 1.35M videos. The fine-tuning accuracy is shown in Table~\ref{tab:kinetics-result} on the Kinetics-400 and Table~\ref{tab:sth-result} on the Sth-Sth V2. We first compare our performance with the original VideoMAE pre-training. We find that for all backbones, our large-scale pre-training obtains better performance than the original VideoMAE pre-trained on the small-scale datasets of Kinetics-400 or Sth-Sth V2. Meanwhile, we see that the performance gap between two pre-training data datasets becomes more evident as the modal capacity scales up. In particular, on the Sth-Sth V2 dataset, our pre-trained ViT-H outperforms the original VideoMAE by 2.0\%. It implies that data scale is also important for video masked autoencoding. Meanwhile, we compare with MAE-ST of IG-uncurated pre-training (1M clips). With the same ViT-L backbone, our pre-trained model outperforms it by 1\% on the Kinetics-400 dataset. This result shows that data quality and diversity might be another important factor. 

\paragraph{Results on model scaling.} We study the performance trend with different model capacities. We compare the fine-tuning performance of pre-trained models with ViT-B, ViT-L, ViT-H, and ViT-g as shown in Table~\ref{tab:kinetics-result} and Table~\ref{tab:sth-result}. Vit-g is the first billion-level model pre-trained in video domain. We obtain consistent performance improvement with increasing model capacity. For all compared pre-training methods, the performance improvement from ViT-B to ViT-L is more obvious, while the improvement from ViT-L to ViT-H is much smaller. 
We further scale up the model capacity to ViT-g architecture. We can still boost the fine-tuning performance further on these two benchmarks with a smaller improvement. We also notice that the performance gap between huge and giant model is very small (0.1\%-0.2\%) in images~\cite{datascaling}. We analyze the performance seems to saturate around 87.0 on the Kinetics-400 and 77.0 on the Sth-Sth V2 for methods without using any extra labeled data. 

\paragraph{Results on progressive training.} We study the influence of the post-pre-training step in our progressive training scheme. To mitigate over-fitting risk and integrate more human supervision into our pre-trained video models, we merge different versions of Kinetics for post-pre-training (intermediate fine-tuning) and evaluate on the Kinetics-400 dataset. The results are reported in Table~\ref{tab:progressive-training}. We observe that the post-pre-training boosts the performance of large-scale pre-trained models for both ViT-H and ViT-g. This result agrees with the findings in image domain~\cite{BEIT,swinv2}. We also apply this technique to VideoMAE V1 pre-trained model and it (ViT-H) achieves worse performance (88.1\% vs. 88.6\%), demonstrating the effectiveness of large-scale unsupervised pre-training. We also compare with the intermediate fine-tuning of MAE-ST and our performance is better by 1.8\% with the same ViT-H backbone. This superior performance might be ascribed to the larger unsupervised pre-training dataset and larger intermediate fine-tuning dataset. We also try this post-pre-training on Sth-Sth V2 dataset but obtain worse performance.

\subsection{Performance on Downstream Tasks}
To demonstrate the generalization ability of our pre-trained models, we transfer their representations to a variety of downstream tasks. In total, we study three kinds of tasks and report performance on ten mainstream benchmarks.

\paragraph{Action classification.}
We compare with previous state-of-the-art methods on four action recognition benchmarks Kinetics-400/600 and Something-Something V1/V2. On the Kinetics datasets, our VideoMAE V2 achieves the best performance among these methods without using extra in-house labeled data, and quite competitive performance to the leading performance MTV~\cite{mtv} trained with extra 60M labeled videos. On the Something-Something datasets, our models significantly outperform the previous best performance, and in particular on Something-Something V1, the performance improvement is above 7\%. The fine-tuning results on UCF101 and HMDB51 are shown in the appendix.

\paragraph{Spatial action detection.}
We perform a comparison with the previous action detection methods on two datasets: AVA and AVA Kinetics. We follow the same detection pipeline with the original VideoMAE~\cite{videomae}. On the AVA dataset, our pre-trained model is significantly better than the previous state-of-the-art action detector of TubeR~\cite{tuber} and outperforms the previous masked modeling methods by 3.1\%. On the AVA-Kinetics, we compare the previous challenge winner methods with the model ensemble, and our single model is better than the best performance by 3.4\%.

\paragraph{Temporal action detection.}
We investigate the transfer performance of our model to the task of temporal action detection. This is an important task yet not to be tested by previous masked pre-training methods. We report performance on two benchmarks: THUMOS14 and FineAction. We use the ActionFormer~\cite{actionformer} detection pipeline as our baseline method and replace its I3D feature with our pre-trained representation. On the THUMOS14 dataset, our model outperforms all previous results even with optical flow as input. On the large-scale FineAction dataset, our model is significantly better than a previous best performance by 5\%.

\section{Conclusion and Discussion}
Building foundation model has turned out to be an effective paradigm to improve the performance of tasks in AI. Simple and scalable algorithms are the core of building powerful foundation models. In this paper, we have presented a simple and efficient way to scale VideoMAE to billion-level model on the million-level pre-training set. Thanks to our core design of dual masking, we are able to successfully the first billion-level video transformer, and demonstrate its effectiveness on a variety of video downstream tasks. Our work shows that video masked autoencoders are general and scalable representation learners for video action understanding. We hope our pre-trained models could provide effective representations for more video understanding tasks in the future.

In spite of these excellent results, challenge still remains. We observe that the performance improvement is smaller when we scale VideoMAE from ViT-H to ViT-g, partially because of performance saturation on these video benchmarks. However, the data scale we have explored for VideoMAE is still several orders of magnitudes smaller than the Image~\cite{swinv2,scalingvit} and NLP~\cite{bert,gpt}. How to train VideoMAE on billions of videos is still extremely challenging for the current software and hardware. We need to come up with more efficient video pre-training framework and hope our work can inspire future work on scaling video pre-training.
	
\paragraph {\bf Acknowledgements.} {This work is supported by the National Key R$\&$D Program of China (No. 2022ZD0160900, No.2022ZD0160100), the National Natural Science Foundation of China (No. 62076119, No. 61921006), Shanghai Committee of Science and Technology (Grant No. 21DZ1100100), the Youth Innovation Promotion Association of Chinese Academy of Sciences (No. 2020355).}

\clearpage
\appendix
\section{Appendix}
%%%%%%%%%%%%%%%%%%%%%%%%%%%%%%%%%%%%%%%%%%%%%%%%%%%%%%%%%%%%%%%%%%
\begin{table*}[htbp]
    \centering
    \renewcommand\arraystretch{1.1}
    \begin{tabular}{c|c|c}
    Stage & VideoMAE V2-giant & Output Size \\
    \shline\hline
    Data & UnlabeldHybrid & $3 \x 16 \x 224 \x 224$ \\
    \hline
    Cube & $\makecell{2 \x 14 \x 14, 1408 \\ \text{stride } 2 \x 14 \x 14}$ & $1408 \x 8 \x 256$ \\
    \hline
    Mask & \makecell{tube masking \\ mask ratio = $\rho$} & $1408 \x  8 \x \lfloor 256 \x (1 - \rho) \rfloor$ \\
    \hline
    Encoder & $\begin{bmatrix} \text{MHA}(1408) \\ \text{MLP}(6144)\end{bmatrix} \x 40$ & $1408 \x 8 \x \lfloor 256 \x (1 - \rho) \rfloor$ \\
    \hline
    Projector & MLP(512) & $512 \x 8 \x \lfloor 256 \x (1 - \rho) \rfloor$ \\
    \hline
    Decoder Mask & \makecell{running cell masking \\ decoder mask ratio = $\rho^d$ \\ concat unmasked learnable tokens} & $512 \x 8 \x \left(\lfloor 256 \x (1 - \rho) \rfloor + \lfloor 256 \x (1 - \rho^d) \rfloor\right)$ \\
    \hline
    Decoder & $\begin{bmatrix} \text{MHA}(512) \\ \text{MLP}(2048)\end{bmatrix} \x 4$ & $512 \x 8 \x \left(\lfloor 256 \x (1 - \rho) \rfloor + \lfloor 256 \x (1 - \rho^d) \rfloor\right)$ \\
    \hline
    Projector & \makecell{discard visible tokens \\ MLP($1176$)} & $1176 \x 8 \x \lfloor 256 \x (1 - \rho^d) \rfloor$ \\
    \hline
    Reshape & from $1176$ to $3\x 2 \x 14 \x 14$ & $3\x 16 \x \lfloor 224(1 - \rho^d) \rfloor \x \lfloor 224(1 - \rho^d) \rfloor$
    \end{tabular}
    \caption{\textbf{Architectures details of VideoMAE V2-g}. The main difference between VideoMAE v2 and VideoMAE v1 is the dual masking design. VideoMAE v2 does not reconstruct the full video clip, while only calculates MSE loss on tokens that are invisible to the encoder.}
    \label{tab:videomae_arch}
\end{table*}
%%%%%%%%%%%%%%%%%%%%%%%%%%%%%%%%%%%%%%%%%%%%%%%%%%%%%%%%%%%%%%%%%%
%%%%%%%%%%%%%%%%%%%%%%%%%%%%%%%%%%%%%%%%%%%%%%%%%%%%%%%%%%%%%
\begin{table}[htbp]
    \centering
    \renewcommand\arraystretch{1.1}
    \begin{tabular}{l|cc}
    Dataset & Size & Source \\
    \shline \hline
    K710 & 658k & YouTube \\
    SSV2 & 169k & Shot from Scripts \\
    AVA & 21k & Movie \\
    WebVid2M & 250k & Internet \\
    self-collected & 250k & Instagram \\
    \hline
    UnlabeledHybrid & 1.348M & Multi-Source
    \end{tabular}
    \caption{\textbf{Components of UnlabeledHybrid}. We build our unlabeled pre-train dataset by collecting clips from multiple resources to ensure the generalization ability of learned models by our VideoMAE V2.}
    \label{tab:unlabeledhybridl}
\end{table}
%%%%%%%%%%%%%%%%%%%%%%%%%%%%%%%%%%%%%%%%%%%%%%%%%%%%%%%%%%%%%
%%%%%%%%%%%%%%%%%%%%%%%%%%%%%%%%%%%%%%%%%%%%%%%%%%%%%%%%%%%%%
\begin{table*}[t]
    \centering
    \renewcommand\arraystretch{1.1}
    \begin{tabular}{l|ccccc}
    Config & \makecell{Kinetics \\ $16 \x 224^2$} & \makecell{Kinetics \\ $64 \x 266^2$} &  \makecell{Sth-Sth \\ $16 \x 224^2$} & \makecell{UCF101 \\ $16 \x 224^2$} & \makecell{HMDB51 \\ $16 \x 224^2$} \\
    \shline
    optimizer & \multicolumn{5}{c}{AdamW} \\
    base learning rate & 1e-5 & 1e-4 & 3e-4 &1e-3&5e-4\\
    weight decay & \multicolumn{5}{c}{0.05} \\
    optimizer momentum &\multicolumn{5}{c}{$\beta_1, \beta_2 = 0.9, 0.999$}\\
    batch size & 32 & 32 & 96 & 24 & 24\\
    learning rate schedule & \multicolumn{5}{c}{cosine decay} \\
    warmup epoch & 0 & 0 & 5 & 5 & 5\\
    epoch & 3 & 5 & 10 & 50 & 15\\
    repeated augmentation & \multicolumn{5}{c}{2}\\
    RandAug & \multicolumn{5}{c}{(0, 0.5)} \\
    label smoothing & \multicolumn{5}{c}{0.1} \\
    mixup & \multicolumn{5}{c}{0.8} \\
    cutmix & \multicolumn{5}{c}{1.0} \\
    drop path & 0.3 & 0.35 & 0.35 & 0.35 & 0.35\\
    flip augmentation & yes & yes & no & yes & yes\\
    augmentation & \multicolumn{5}{c}{MultiScaleCrop} \\
    dropout & \multicolumn{5}{c}{0.5} \\
    layer-wise lr decay & \multicolumn{5}{c}{0.9} \\
    clip grading & \multicolumn{5}{c}{None}
    \end{tabular}
    \caption{Action classification setting in specific fine-tuning stage.}
    \label{tab:specific_finetune}
\end{table*}
%%%%%%%%%%%%%%%%%%%%%%%%%%%%%%%%%%%%%%%%%%%%%%%%%%%%%%%%%%%%%
%%%%%%%%%%%%%%%%%%%%%%%%%%%%%%%%%%%%%%%%%%%%%%%%%%%%%%%%%%%%%
\begin{table}[htbp]
    \centering
    \begin{tabular}{l|c}
    Config & Value \\
    \shline
    mask ratio & 0.9 \\
    decoder mask ratio & 0.5 \\
    optimizer & AdamW\cite{adamw} \\
    base learning rate & 1.5e-4 \\
    weight decay & 0.05 \\
    optimizer momentum & $\beta_1, \beta_2 = 0.9, 0.95$ \\
    batch size & 8192 \\
    learning rate schedule & cosine decay \\
    warmup epoch & 120 \\
    epoch & 1200 \\
    repeated augmentation & 4 \\
    flip augmentation & no \\
    augmentation & MultiScaleCrop \\
    clip grading & 0.02
    \end{tabular}
    \caption{\textbf{Pre-training setting}, where batch size includes the additional views produced by repeated augmentation and epochs refers to the total number of times the data is sampled.}
    \label{tab:pretrain_set}
\end{table}
%%%%%%%%%%%%%%%%%%%%%%%%%%%%%%%%%%%%%%%%%%%%%%%%%%%%%%%%%%%%%
%%%%%%%%%%%%%%%%%%%%%%%%%%%%%%%%%%%%%%%%%%%%%%%%%%%%%%%%%%%%%
\begin{table}[htbp]
    \centering
    \begin{tabular}{l|c}
    Config & Value \\
    \shline
    optimizer & AdamW\cite{adamw} \\
    base learning rate & 1e-3 \\
    weight decay & 0.05 (H), 0.1 (g) \\
    optimizer momentum & $\beta_1, \beta_2 = 0.9, 0.999$\cite{iGPT20} \\
    batch size & 128 \\
    learning rate schedule & cosine decay\cite{loshchilov2016sgdr} \\
    warmup epoch & 5 \\
    epoch & 40 (H), 35 (g) \\
    repeated augmentation\cite{hoffer2020augment} & 2 \\
    RandAug\cite{cubuk2020randaugment} & (0, 0.5) \\
    label smoothing\cite{szegedy2016rethinking} & 0.1 \\
    mixup\cite{zhang2017mixup} & 0.8 \\
    cutmix\cite{yun2019cutmix} & 1.0 \\
    drop path & 0.2 (H), 0.3 (g) \\
    flip augmentation & yes \\
    augmentation & MultiScaleCrop \\
    dropout & 0.5 (H), None (g) \\
    layer-wise lr decay\cite{BEIT} & 0.8 (H), 0.9 (g) \\
    clip grading & None (H), 5.0 (g)
    \end{tabular}
    \caption{Post-pre-training setting.}
    \label{tab:post_pretrain_set}
\end{table}
%%%%%%%%%%%%%%%%%%%%%%%%%%%%%%%%%%%%%%%%%%%%%%%%%%%%%%%%%%%%%
%%%%%%%%%%%%%%%%%%%%%%%%%%%%%%%%%%%%%%%%%%%%%%%%%%%%%%%%%%%%%
\begin{table}[htbp]
    \centering
    \renewcommand\arraystretch{1.1}
    \begin{tabular}{l|cc}
    Config & AVA 2.2 & AVA-Kinetics \\
    \shline
    optimizer & \multicolumn{2}{c}{AdamW} \\
    base learning rate & \multicolumn{2}{c}{3e-4} \\
    weight decay & \multicolumn{2}{c}{0.05} \\
    optimizer momentum & \multicolumn{2}{c}{$\beta_1, \beta_2 = 0.9, 0.999$} \\
    batch size & \multicolumn{2}{c}{128} \\
    learning rate schedule & \multicolumn{2}{c}{cosine decay} \\
    warmup epoch & \multicolumn{2}{c}{2} \\
    epoch & \multicolumn{2}{c}{10} \\
    repeated augmentation & \multicolumn{2}{c}{no} \\
    drop path & \multicolumn{2}{c}{0.3} \\
    flip augmentation & \multicolumn{2}{c}{yes} \\
    layer-wise lr decay & \multicolumn{2}{c}{0.9} \\
    clip grading & \multicolumn{2}{c}{None} \\
    \end{tabular}
    \caption{Hyper-parameter settings of action detection.}
    \label{tab:ava_finetune}
\end{table}
%%%%%%%%%%%%%%%%%%%%%%%%%%%%%%%%%%%%%%%%%%%%%%%%%%%%%%%%%%%%%
%%%%%%%%%%%%%%%%%%%%%%%%%%%%%%%%%%%%%%%%%%%%%%%%%%%%%%%%%%%%%%
\begin{table}[htbp]
\centering
\renewcommand\arraystretch{1.0}
\begin{tabular}{l|c|c|c}
        \toprule
	\bfseries Method &\bfseries Modality & \bfseries UCF101 & \bfseries HMDB51 \\ 
	\midrule
	OPN~\cite{opn} &  V & 59.6 & 23.8 \\
	VCOP~\cite{vcop} &  V & 72.4 & 30.9 \\
	SpeedNet~\cite{speedNet} & V &  81.1 & 48.8  \\
	VTHCL~\cite{vthcl} & V & 82.1 & 49.2  \\  
        Pace~\cite{wang2020self}& V  & 77.1 & 36.6  \\
        MemDPC~\cite{memdpc} & V & 86.1 & 54.5 \\
	CoCLR~\cite{han2020coclr}  & V & 87.9 & 54.6   \\
	RSPNet~\cite{rspnet} &V  & 93.7 & 64.7 \\
	VideoMoCo~\cite{pan2021videomoco} & V & 78.7 & 49.2  \\
	Vi$^2$CLR~\cite{Diba_2021_ICCV}  & V & 89.1 & 55.7  \\
	CVRL~\cite{cvrl} & V & 94.4 & 70.6  \\ 
	CORP$_{f}$~\cite{Hu_2021_ICCV} & V & 93.5 & 68.0  \\ 
	$\rho$SimCLR$_{\rho=2}$~\cite{large}  &V  & 88.9 & N/A  \\
	$\rho$SwAV$_{\rho=2}$~\cite{large} &V   & 87.3 & N/A  \\
	$\rho$MoCo$_{\rho=2}$~\cite{large} & V  & 91.0 & N/A  \\
	$\rho$BYOL$_{\rho=4}$~\cite{large}  &  V & 94.2 & 72.1  \\
	\midrule
	MIL-NCE~\cite{milnce}  & V+T & 91.3 & 61.0 \\
	MMV~\cite{mmv} &  V+A+T & 92.5 & 69.6 \\
	CPD~\cite{cpd}  & V+T & 92.8 & 63.8  \\
        ELO~\cite{elo} & V+A  & 93.8 & 67.4   \\
	XDC~\cite{xdc} &  V+A & 94.2  & 67.1 \\ 
	GDT~\cite{gdt} & V+A & 95.2 & 72.8  \\
	\midrule
	VideoMAE V1 & V & 96.1 & 73.3 \\
        \textbf{VideoMAE V2} & V & \textbf{99.6} & \textbf{88.1} \\
        \bottomrule
\end{tabular}
        %\vspace{-0.5em}
	\caption{\textbf{Comparison with the state-of-the-art methods on UCF101 and HMDB51.} `V' refers to visual, `A' is audio, `T' is text narration. ``N/A'' indicates the numbers are not available.} \label{tab:ucf_hmdb_supp}
	\vspace{-1em}
\end{table}
%%%%%%%%%%%%%%%%%%%%%%%%%%%%%%%%%%%%%%%%%%%%%%%%%%%%%%%%%%%%%%
\begin{table}[htbp]
\centering
   \resizebox{0.99\linewidth}{!}{ 
   \begin{tabular}{lcccc}
				\toprule
				Method 																			 & Top 1                & Top 5         & Views 	& TFLOPs \\
				\midrule
				I3D NL~\cite{nonlocal}										 & 77.7                  & 93.3         		 &  $10 \times 3$ & 10.77     \\
                TDN~\cite{tdn} & 79.4 & 94.4 &  $10 \times 3$ &  5.94 \\
				SlowFast R101-NL~\cite{Slowfast}       		&  79.8                 &  93.9                   & $10 \times 3$    & 7.02   \\  
				TimeSformer-L~\cite{timsformer} & 80.7				& 94.7					& $1 \times 3$ & 7.14 \\
                MTV-B~\cite{mtv} & 81.8 & 95.0 	& $4 \times 3$ & 4.79 \\ %
                Video Swin~\cite{videoswin} & 83.1 & 95.9 & $4 \times 3$ & 7.25 \\
				MViT-B~\cite{mvit}					  	& 81.2				& 95.1					& $3 \times 3$ & 4.10 \\
				ViViT-L FE~\cite{vivit} & 81.7 	&  93.8  & $1 \times 3$ &  11.94 \\  %
                MViTv2-B~\cite{mvitv2} & 82.9 & 95.7 & $1 \times 5$ & 1.13  \\
                MViTv2-L (312p)~\cite{mvitv2} & 86.1 & 97.0 & $3 \times 5$ & 42.42 \\
                MaskFeat~\cite{maskedfeat} & 85.1 & 96.6 & $1 \times 10$ & 3.78 \\
                MaskFeat (352p)~\cite{maskedfeat}  & 87.0 & 97.4 & $4 \times 3$ & 45.48 \\
                MAE-ST~\cite{MAE-ST} & 86.8 & 97.2 &  $7 \times 3$ &  25.05 \\
                VideoMAE~\cite{videomae} & 86.6 & 97.1  & $5 \times 3$ & 17.88 \\
                VideoMAE (320p) ~\cite{videomae} & 87.4 & 97.6  & $4 \times 3$ & 88.76 \\
                {\bf VideoMAE V2-H} & {88.6} & 97.9 & $5 \times 3$ & 17.88 \\
                {\bf VideoMAE V2-g} & 88.5 & {98.1} & $5 \times 3$ & 38.16 \\
                {\bf VideoMAE V2-H ($64\x 288^2$)} & 89.8 & 98.3 & $4 \times 3$ & 153.34 \\
                {\bf VideoMAE V2-g ($64\x 266^2$)} & \textbf{90.0} & {\bf 98.4} &  $2 \times 3$ & 160.30 \\
                
				\midrule
				\multicolumn{4}{l}{\textit{Methods using in-house labeled data}}                                \\ 
				\textcolor{gray}{CoVeR (JFT-3B)~\cite{zhang2021}} & \textcolor{gray}{87.2} & \textcolor{gray}{--} & \textcolor{gray}{$1 \times 3$} & \textcolor{gray}{-} \\
				\textcolor{gray}{MTV-H (WTS)~\cite{mtv}} &  \textcolor{gray}{89.1} &  \textcolor{gray}{98.2} 	& \textcolor{gray}{$4 \times 3$} & \textcolor{gray}{44.47} \\ %
				\textcolor{gray}{\textbf{MTV-H} (WTS $280^2$)~\cite{mtv}} 	&  \textcolor{gray}{\textbf{89.9}} &  \textcolor{gray}{98.3} 	& \textcolor{gray}{$4 \times 3$} & \textcolor{gray}{73.57} \\ %
				\bottomrule
			\end{tabular}
   }
   \caption{{\bf Results on the Kinetics-400 dataset.} We report the performance of our pre-trained model with larger input resolution and more frames.}
   \label{tab:k400}
\end{table}
%%%%%%%%%%%%%%%%%%%%%%%%%%%%%%%%%%%%%%%%%%%%%%%%%%%%%%%%%%%%%%
%%%%%%%%%%%%%%%%%%%%%%%%%%%%%%%%%%%%%%%%%%%%%%%%%%%%%%%%%%%%%%
\begin{table}[htbp]
\centering
   \resizebox{0.99\linewidth}{!}{ 
   			\begin{tabular}{lcccc}
				\toprule
				Method 																			 & Top 1                & Top 5         & Views 	& TFLOPs \\
				\midrule
				
				SlowFast R101-NL~\cite{Slowfast}       		&  81.8                &  95.1                  & $10 \times 3$    & 7.02   \\ 
				TimeSformer-L~\cite{timsformer} & 82.2				& 95.6					& $1 \times 3$ & 7.14 \\
                MTV-B~\cite{mtv} & 83.6 & 96.1 	& $4 \times 3$ & 4.79 \\ 
				MViT-B~\cite{mvit}					  	& 83.8			& 96.3				& $3 \times 3$ & 4.10 \\
				ViViT-L FE~\cite{vivit} & 82.9 	&  94.6  & $1 \times 3$ &  11.94 \\  
                MViTv2-B~\cite{mvitv2} & 85.5 & 97.2 & $1 \times 5$ & 1.03  \\
                MViTv2-L (352p)~\cite{mvitv2} & 87.9 & 97.9 & $3 \times 4$ & 45.48 \\
                MaskFeat~\cite{maskedfeat} & 86.4 & 97.4 & $1 \times 10$ & 3.77 \\
                MaskFeat (312p)~\cite{maskedfeat} & 88.3 & 98.0 & $3 \times 4$ & 33.94 \\
                {\bf VideoMAE V2-H} & 88.3 & 98.1 & $5 \times 3$ & 17.88 \\
                {\bf VideoMAE V2-g} & 88.8 & 98.2 & $5 \times 3$ & 38.16 \\
                {\bf VideoMAE V2-H ($32\x 384^2$)} &  89.6 & 98.4 & $4 \times 3$ & 184.24\\
                {\bf VideoMAE V2-g ($64\x 266^2$)} &\textbf{ 89.9} & \textbf{98.5} & $2 \times 3$ & 160.30\\
				\midrule
				\multicolumn{4}{l}{\textit{Methods using in-house labeled data}}                                \\ 
				\textcolor{gray}{CoVeR (JFT-3B)~\cite{zhang2021}} & \textcolor{gray}{87.9} & \textcolor{gray}{97.8} & \textcolor{gray}{$1 \times 3$} & \textcolor{gray}{-}\\
				\textcolor{gray}{MTV-H (WTS)~\cite{mtv}} 														&  \textcolor{gray}{89.6} &  \textcolor{gray}{98.3} 	& \textcolor{gray}{$4 \times 3$} & \textcolor{gray}{44.47} \\ %
				\textcolor{gray}{\textbf{MTV-H} (WTS $280^2$)~\cite{mtv}} 														&   \textcolor{gray}{\textbf{90.3}} &   \textcolor{gray}{\textbf{98.5}} 	&  \textcolor{gray}{$4 \times 3$} &  \textcolor{gray}{73.57} \\ %
				\bottomrule
			\end{tabular}
   }
   \caption{{\bf Results on the Kinetics-600 dataset.} We report the performance of our pre-trained model with larger input resolution and more frames.}
   \label{tab:k600}
\end{table}
%%%%%%%%%%%%%%%%%%%%%%%%%%%%%%%%%%%%%%%%%%%%%%%%%%%%%%%%%%%%%%
%%%%%%%%%%%%%%%%%%%%%%%%%%%%%%%%%%%%%%%%%%%%%%%%%%%%%%%%%%%%%
\begin{table}[htbp]
    \centering
    \begin{tabular}{l|c}
    Config & Value \\
    \shline
    optimizer & AdamW\cite{adamw} \\
    base learning rate & 1e-3~(K710),~5e-4~(SSv2) \\
    weight decay & 0.05 \\
    optimizer momentum & $\beta_1, \beta_2 = 0.9, 0.999$\cite{iGPT20} \\
    batch size & 1024~(K710),~512~(SSv2) \\
    learning rate schedule & cosine decay\cite{loshchilov2016sgdr} \\
    warmup epoch & 5 \\
    epoch & 100 \\
    RandAug\cite{cubuk2020randaugment} & (0, 0.5) \\
    mixup\cite{zhang2017mixup} & 0.8 \\
    cutmix\cite{yun2019cutmix} & 1.0 \\
    drop path & 0.1 \\
    flip augmentation & yes \\
    augmentation & MultiScaleCrop \\
    dropout & None \\
    layer-wise lr decay\cite{BEIT} & 0.75 \\
    clip grading & 1.0 \\
    temperature & 3.0
    \end{tabular}
    \caption{Knowledge distilling setting.}
    \label{tab:kd_setting}
\end{table}
%%%%%%%%%%%%%%%%%%%%%%%%%%%%%%%%%%%%%%%%%%%%%%%%%%%%%%%%%%%%%
\begin{table}[htbp]
    \centering
    \begin{tabular}{c|ccc}
    \toprule
    Model &  K400 & K600 & SSv2 \\
    \midrule
    VideoMAE-B & 81.5 & N/A & 70.8 \\
    VideoMAE V2-g & 88.5 & 88.8 & 77.0 \\
    \midrule
    Distilled ViT-B & 87.1 & 87.4 & 75.0 \\
    \bottomrule
    \end{tabular}
    \vspace{-0.5em}
    \caption{The performance of distilled ViT-B models on the datasets of Kinetics400, Kinetics600, and Something-Something V2.}
    \label{tab:distil_result}
\end{table}
%%%%%%%%%%%%%%%%%%%%%%%%%%%%%%%%%%%%%%%%%%%%%%%%%%%%%%%%%%%%%

In this supplementary material, we provide more details of our VideoMAE V2 and present more experiment results. Specifically, we give a detailed description of the architectures of our VideoMAE V2 in Section~\ref{sec:arc}. Then, we present the details on building our pre-training datasets in Section~\ref{sec:data}. After this, we provide more implementation details in our experiments in Section~\ref{sec:details}. Finally, we give more results and analysis on our VideoMAE V2 in Section~\ref{sec:result}.

\section{Model Architecture}
\label{sec:arc}
We build the encoder and decoder in our VideoMAE V2 by using the vanilla ViT backbone with joint space-time attention. To ensure efficient computation, our decoder does not get larger as the encoder scales up, but always stays at the size of 4 layers and 512 channels. We show the architectures of VideoMAE V2 in Table~\ref{tab:videomae_arch}, taking ViT-giant as an example.

\section{Datasets}
\label{sec:data}
\subsection{UnlabeledHybrid}
Our UnlabeledHybrid dataset is a hybrid dataset consisting of Kinetics~\cite{kinetics}, Something-Something~\cite{sth}, AVA~\cite{ava}, WebVid2M~\cite{webvid2m}, and our self-collected Instagram dataset. When training VideoMAE V2, the sampling stride $\tau$ is 2 on Something-Something, and 4 on the other datasets. The detailed components of UnlabeledHybrid are shown in Table~\ref{tab:unlabeledhybridl}. We now specify the handling of each dataset.

\paragraph{Kinetics.} Videos in Kinetics are from YouTube. We adapt the same method with \cite{anonymous2023uniformerv} to make a mixed kinetics dataset. Kinetics has three versions, Kinetics-400/600/700, based on the number of human action categories. We merge the training set and validation set of the three versions, then remove the duplicated videos according to YouTube IDs, and finally delete the validation videos that existed in the training set. As some videos have different category names in different versions of Kinetics, we also group them together, resulting in a Kinetics dataset with 710 categories, termed Kinetics-710 (K710) or \textit{LabeledHybrid}. K710 contains 658k training videos and 67k validation videos.

\paragraph{Something-Something.} Videos in Something-Something are shot from video scripts, usually from a first-person perspective. We choose Something-Something V2 (SSV2) as the part of UnlabeledHbyrid dataset. SSV2 is a motion-centric dataset containing 169k training videos and 25k validation videos.

\paragraph{AVA.} Videos in AVA are movie clips, ranging from the 15th to the 30th minute of each movie. We always cut the 15-minute movie clips from the AVA training set by 300 frames, resulting in 21k video clips.

\paragraph{WebVid2M.} Videos in WebVid2M are scraped from the internet. We randomly pick 250k training videos from the original datasets.

\paragraph{Self-collected Instagram dataset.} We used thousands of category tags from the already publicly available dataset as query phrases to scrape million of videos from Instagram. The average duration of the videos is 34 seconds. We also randomly pick 250k videos from the dataset.

\subsection{LabledHybrid}
We build the labeled datasets for our VideoMAE post-pre-training by taking the union of different versions of Kinetics dataset. The construction details is following the UniformerV2~\cite{uniformerv2} and more details could be referred to the original paper.

\section{Implementation Details}
\label{sec:details}
In this section, we will describe the implementation details in the three stages of progressive training: \textit{pre-training}, \textit{post-pre-training}, and \textit{specific fine-tuning}.

\subsection{Pre-training}
We pre-train VideoMAE V2, both ViT-huge and ViT-giant, 1200 epochs on the UnlabeledHybrid dataset with 64 80G-A100 GPUs. Besides the dual masking core design of VideoMAE V2, we also adapt mix-precision training and checkpointing at the engineering level to speed up pre-training. To avoid the potential precision overflow risk during model pre-training, we train the encoder with FP-16 mixed precision and the decoder with FP32 precision. We adapt repeated augmentation to reduce the video loading overhead. The learning rate is scaled linearly according to the total batch size, \ie $\text{lr} = \text{base\_lr} \times  \text{batch\_size}~ / ~256$. The detailed pre-training setting is shown in Table~\ref{tab:pretrain_set}.

\subsection{Post-pre-training}
In the supervised \textit{post-pre-training} stage, we fine-tune the pre-trained encoder on \textit{LabeledHybrid} (K710). When training ViT-giant, we found that the dropout layer before the classification head has little positive effect on preventing overfitting, so the dropout layer was removed and the drop path rate was increased slightly. The clip grading stabilizes the optimization of large models in the early stages of fine-tuning to some extent, and it is advisable to adjust the value of the clip grading with the batch size changing. The choice of layer decay matters. A smaller layer decay better maintains the pre-training effect, but may not provide enough space for improvement in the later stages of fine-tuning. A relatively large layer decay is recommended when the model is well pre-trained, i.e. when it exhibits smaller gradients in the shallow layers and bigger gradients in the deep layers at the early stages of fine-tuning. The detailed settings are shown in Table~\ref{tab:post_pretrain_set}. Notably, this setting also works for fine-tuning directly on the kinetics dataset.

\subsection{Specific fine-tuning}
After the post-pre-training stage, we perform the \textit{specific fine-tuning} stage to get the specific models on action classification, action detection, and temporal action detection.

\subsubsection{Action classification}
We test the performance of the specific models for action classification on Kinetics~\cite{kinetics}, Something-Something~\cite{sth}, UCF101~\cite{ucf} and HMDB51~\cite{hmdb} with regular $16\x 224^2$ inputs. Further, we also test the performance of the model on Kinetics~\cite{kinetics} with larger input shapes $64 \x 266^2$. The detailed fine-tuning setting of VideoMAE V2-g can be seen in Table~\ref{tab:specific_finetune}. At the specific fine-tuning stage, increasing the dropout and drop path can reduce the risk of overfitting to a certain extent, and the optimization of the model is more stable after the supervised post-pre-training, so clip grading is not necessary.

\subsubsection{Action detection}
We follow the training pipeline of the original VideoMAE \ie person detection + action classification. We adapt only two data augmentations, random scale cropping, and random horizontal flipping. When training, we use the ground-truth person boxes, while in testing, we use the person boxes detected by AIA\cite{tang2020asynchronous}. More settings see in Table~\ref{tab:ava_finetune}.

\subsubsection{Temporal action detection}
We take the model trained on the \textit{LabeledHybrid} dataset as the backbone network and test its generalization performance on the temporal action detection task following the architecture of ActionFormer\cite{actionformer} on THUMOS14\cite{thumos} and FineAction\cite{fineaction}. When training, we use Adam\cite{kingma2014adam} with warm-up and fix the maximum input sequence length. As for inference, we use Soft-NMS\cite{bodla2017softnms} on the result action candidates to remove the highly overlap proposals and obtain the final result.

\section{More Results}
\label{sec:result}

\paragraph{\bf More results and analysis.} We report more result comparisons on Kinetics with the larger input size in Table~\ref{tab:k400} and Table~\ref{tab:k600}. We also add the results on UCF101~\cite{ucf} and HMDB51~\cite{hmdb} in Table~\ref{tab:ucf_hmdb_supp}. From these results, we see that our model can further improve the recognition results by using larger input. Meanwhile, on the smaller benchmarks of UCF101 and HMDB51, our model obtains state-of-the-art performance, which is much better than the VideoMAE V1.

\paragraph{\bf Distillation results.} Using the procedure of \cite{beyer2022knowledge}, we are able to compress VideoMAE V2-g into a mush smaller ViT-B. Specifically, we initialize the student model with the VideoMAE V2-B weights after the post-pre-training. Then, we conduct the distillation on K710 (or SSv2) dataset for 100 epochs, with the goal of minimizing the KL divergence between the student model's logits and those of the teacher model. Detailed settings see in Table~\ref{tab:kd_setting}. Our evaluation of the distilled ViT-B model is based on its performance on the K400, K600, and SSv2, as shown in Table~\ref{tab:distil_result}. From these results, we see that our distilled ViT-B model achieves much better performance than the original VideoMAE ViT-B models. We hope our distilled ViT-B model can serve as an efficient foundation model for downstream tasks.

%%%%%%%%% REFERENCES
{\small
\bibliographystyle{ieee_fullname}
\bibliography{egbib}
}

\end{document}